\documentclass[10pt,twocolumn,letterpaper]{article}

\usepackage{iccv}
\usepackage{times}
\usepackage{epsfig}
\usepackage{graphicx}
\usepackage{amsmath}
\usepackage{amssymb}

\usepackage{placeins}
\usepackage{enumitem}
\usepackage{multirow}

\usepackage{amsfonts}
\usepackage{amsthm}

\usepackage{amsmath}

\newcommand{\fig}[1]{Figure~\ref{fig:#1}}
\newcommand{\sect}[1]{Section~\ref{sect:#1}}
\newcommand{\tab}[1]{Table~\ref{tab:#1}}

\newcommand{\eq}[1]{(\ref{eq:#1})}

\usepackage[pagebackref=true,breaklinks=true,letterpaper=true,colorlinks,bookmarks=false]{hyperref}

\iccvfinalcopy 

\def\iccvPaperID{3300} 

\ificcvfinal\fi
\begin{document}

\title{Unsupervised Neural Quantization for Compressed-Domain Similarity Search}


\author{
   Stanislav Morozov \\
   Yandex, \\
   Lomonosov Moscow State University \\
   {\tt\small stanis-morozov@yandex.ru}
   \and
   Artem Babenko \\
   Yandex,\\
   National Research University \\Higher School of Economics \\
   {\tt\small artem.babenko@phystech.edu}
}

\maketitle

\begin{abstract}
We tackle the problem of unsupervised visual descriptors compression, which is a key ingredient of large-scale image retrieval systems. While the deep learning machinery has benefited literally all computer vision pipelines, the existing state-of-the-art compression methods employ shallow architectures, and we aim to close this gap by our paper. In more detail, we introduce a DNN architecture for the unsupervised compressed-domain retrieval, based on multi-codebook quantization. The proposed architecture is designed to incorporate both fast data encoding and efficient distances computation via lookup tables. We demonstrate the exceptional advantage of our scheme over existing quantization approaches on several datasets of visual descriptors via outperforming the previous state-of-the-art by a large margin.
\end{abstract}

\section{Introduction}
\label{sect:intro}

Unsupervised compression of high-dimensional visual descriptors has a long history in the computer vision community\cite{semhash,PQ}. Nowadays, the development of effective compact representations becomes even more crucial for the scalability of modern search engines, given the enormous amount of visual data in the Web.

Currently the dominant unsupervised compression methods\cite{PQ,OPQ,Norouzi13,AQ,CQ,LSQ,LSQ++} belong to the multi-codebook quantization (MCQ) paradigm. In this paradigm, descriptors are effectively approximated by a sum or a concatenation of a few $codeword$ vectors, coming from several disjoint codebooks. Such a simple form of approximation enables the efficient computation of distances from uncompressed queries to compressed database vectors via the usage of lookup tables. While originally appeared for the image retrieval problem, the quantization methods are extensively used to increase the efficiency in a wide range of tasks, e.g. CNN compression\cite{GongLYB14,ZhangL18} or localization\cite{LynenSBHPS15}. Indeed, the development of more advanced quantization methods remains an important research direction as they would benefit a whole range of large-scale computer vision applications.

Despite the ubiquitous use of deep architectures in different areas of computer vision, the unsupervised quantization for the compressed-domain retrieval, which we tackle in this paper, did not benefit from their power yet. While several recent works investigate the usage of deep architectures for the supervised MCQ scenario\cite{YuYFJ18,JainZPG17,dpq}, the state-of-the-art unsupervised methods\cite{LSQ,CompQ,LSQ++} remain shallow. Moreover, the recent work\cite{SablayrollesDUJ17} has shown that even for the supervised compression problem, the usage of unsupervised MCQ outperforms several strong supervised baselines. To the best of our knowledge, at the moment it is not clear if the deep learning machinery can benefit the unsupervised quantization approaches. This is the central question we aim to answer in our paper.

As the main novelty, we introduce a new \textbf{Unsupervised Neural Quantization (UNQ)} method, which learns a nonlinear multi-codebook quantization model, trainable via SGD. Our model is partially inspired by the ideas from the recent works on generative modeling with discrete hidden variables\cite{gumbel,concrete,vq_vae}, which, as we show, appear to be a natural fit for the compressed-domain retrieval problem. In its essence, UNQ works via embedding both codewords and data vectors into a common learnable vector space, where efficient nearest neighbor retrieval is possible. As we show in the experimental section, the non-linear nature of our model allows increasing the retrieval accuracy, compared to the existing shallow competitors.

Overall, the main contributions of our paper can be summarized as follows:
\begin{enumerate}
\item We propose a new method for the unsupervised quantization of visual descriptors. To the best of our knowledge, our method is the first successful case of the usage of deep architectures for the unsupervised MCQ for compressed-domain retrieval. 

\item With the extensive experimental evaluation, we show that the proposed method outperforms the existing techniques in terms of retrieval performance. For most operating points our method provides a new state-of-the-art on two common benchmarks.

\item The Pytorch implementation of the proposed method is available online\footnote{\texttt{https://github.com/stanis-morozov/unq}}. 
\end{enumerate}

The rest of the paper is organized as follows. In \sect{related} we review the existing unsupervised quantization approaches. The proposed Unsupervised Neural Quantization model is described in \sect{method} and experimentally evaluated in \sect{experiments}. \sect{conclusion} concludes the paper.

\section{Related work}
\label{sect:related}

In this section we briefly review the main ideas from the previous works that are relevant for our method.

\textbf{High-dimensional data compression.} The existing methods for the high-dimensional data compression mostly fall into two separate lines of research. The first family\cite{semhash,itq,spherhash} includes binary hashing methods, which map the original vectors into the Hamming space such that nearby vectors are mapped into hashes with small Hamming distances. The practical advantage of binary hashing is that it heavily benefits from the efficient binary computations in modern CPU architectures. The second family of methods generalizes the idea of vector quantization, and we refer to these methods as multi-codebook quantization (MCQ). MCQ methods typically do not involve the information loss on the query side, hence they typically outperform the binary hashing methods by a large margin. The state-of-the-art compression accuracy is currently achieved by the recent MCQ methods\cite{CompQ,LSQ++} and in this paper we aim to improve their quality even further via the power of deep architectures.

\textbf{Product quantization (PQ)}\cite{PQ} is a pioneering method from the MCQ family, which inspired further research on this subject. PQ encodes each vector $x \in \mathbf{R}^D$ as a concatenation of $M$ codewords from $M$ $\frac{D}{M}$\nobreakdash-dimensional codebooks $C_1,\ldots,C_M$, each containing $K$ codewords. In other words, PQ decomposes a vector into $M$ separate subvectors and applies vector quantization (VQ) to each subvector, while using a separate codebook. As a result each vector $x$ is encoded by a tuple of codewords indices $[i_1,\ldots,i_M]$ and approximated by $x \approx [c_{1i_1},\ldots, c_{Mi_M}]$.
Fast Euclidean distance computation becomes possible via efficient ADC procedure\cite{PQ} using lookup tables:

\begin{gather}
\label{eq:adc}
\|q - x\|^2 \approx \|q - [c_{1i_1},\ldots, c_{Mi_M}]\|^2 = \\
\sum\limits_{m=1}^{M}{\|q_m - c_{mi_m}\|}^2 \nonumber
\end{gather}

where $q_m$ --- $m$th subvector of a query $q$. This sum can be calculated in $M$ additions and lookups given that distances from query subvectors to codewords are precomputed.

From the geometry viewpoint, PQ effectively partitions the original vector space into $K^M$ cells, each being a Cartesian product of $M$ lower-dimensional cells. Such product-based approximation works better if the $\frac{D}{M}$-dimensional components of vectors have independent distributions. The degree of dependence is affected by the choice of the splitting, and can be further improved by orthogonal transformation applied to vectors as preprocessing. Two subsequent works have therefore looked into finding an optimal transformation \cite{OPQ,Norouzi13}. The modification of PQ corresponding to such pre-processing transformation is referred below as Optimized Product Quantization (OPQ).

\begin{figure*}[t]
    \centering
    \includegraphics[width=520px]{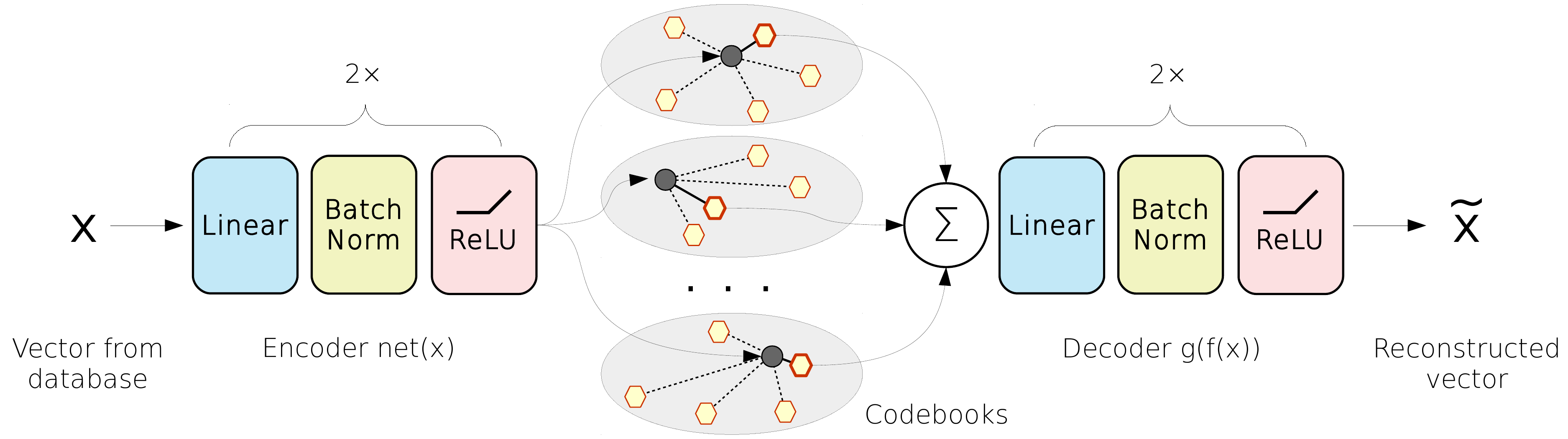}
    \vspace{1mm}
    \caption{The proposed Unsupervised Neural Quantization model architecture. The encoder(left) maps data vector into a product of learned spaces (middle), selects codewords and decodes them back into the original vector space (right). The grey ellipses represent the codebook spaces and the orange hexagons denote the codewords in those codebooks.}
    \label{fig:arch}
\end{figure*}

\textbf{Non-orthogonal quantizations.} Several works \cite{RVQ, AQ, CQ, LSQ, CompQ, LSQ++} generalize the idea of Product Quantization by approximating each vector by a sum of $M$ codewords instead of concatenation. In this case, the ADC procedure is still efficient while the approximation accuracy is increased. 

The first approach, Residual Vector Quantization \cite{RVQ}, quantizes original vectors, and then iteratively quantizes the approximation residuals from the previous iteration. Another approach, Additive Quantization (AQ) \cite{AQ}, is the most general as it does not impose any constraints on the codewords from the different codebooks. Usually, AQ provides the smallest compression errors, however, it is much slower than other methods, especially for large $M$. Composite Quantization (CQ) \cite{CQ} learns codebooks with a fixed value of scalar product between the codewords from different codebooks. Several recent works\cite{LSQ,LSQ++,CompQ} elaborate the idea of Additive Quantization, proposing the more effective procedure for codebooks learning. Currently, state-of-the-art compression accuracy is achieved by the LSQ method\cite{LSQ++}.
We present the qualitative comparison of the existing MCQ methods with the open-source implementations as well as the proposed UNQ method in Table $1$.

\begin{table}
\addtolength{\tabcolsep}{1.5pt}
\renewcommand\arraystretch{1.1}
\begin{tabular}{|c|c|c|c|}
\hline
Method & {(O)PQ} & {AQ/LSQ} & {UNQ} \\
\hline
Compression Quality & Medium & High & High\\
\hline
Encoding complexity & Low & High & Low \\
\hline
Learning complexity & Low & High & High \\
\hline
\end{tabular}
\label{tab:mcq_comparison}
\caption{The qualitative comparison of Unsupervised Neural Quantization (UNQ) with the existing quantization methods.}
\end{table}

\textbf{Compression with DNN.} Several recent works\cite{YuYFJ18,JainZPG17,dpq} investigate the usage of deep architectures for multi-codebook quantization in the supervised compression scenario. In contrast, we tackle the more challenging unsupervised setup, where only shallow quantization methods are currently in use. To the best of our knowledge, there is only one recent paper\cite{catalyst} that employs a deep architecture for the unsupervised compression problem, but it does not work in the MCQ paradigm. Instead, \cite{catalyst} performs neighborhood-preserving mapping to a sphere with an additional "spreading" regularizer that enforces the uniform distribution of mapped data points. Then \cite{catalyst} uses the fixed predefined lattices to quantize the data vectors. In our experiments, we demonstrate the advantage of UNQ over \cite{catalyst} in most operating points.

At the same time, several recent studies on generative modeling developed efficient ways to learn discrete representations with deep neural networks. One branch of such methods relies on continuous noisy relaxations of the discrete variables that can be trained by backpropagation \cite{gumbel,concrete,rebar}. Another popular approach to learning discrete variables is featured in Vector-Quantized Variational Autoencoder model\cite{vq_vae,ef_vq_vae}. Instead of continuous relaxation, this approach uses straight-through gradient estimation to propagate gradient through discrete variables. To the best of our knowledge, our work is the first that experimentally demonstrates that with the appropriate training protocol, the discrete hidden variables can be successfully used for compressed-domain retrieval in the unsupervised scenario. While we acknowledge the existence of the recent concurrent preprint\cite{pqvae}, which studies the close ideas, its experimental evaluation shows the performance of their approach to be on par with binary hashing methods, which are improper baselines. In contrast, our UNQ approach substantially outperforms the current state-of-the-art, as will be shown in experiments. While in this paper our approach is used only for compression of precomputed descriptors, the proposed architecture can be combined with existing self-supervised methods\cite{deepcluster} for end-to-end unsupervised image compression.


\section{Unsupervised Neural Quantization}
\label{sect:method}
%
%

We now introduce notation and discuss the proposed UNQ method in detail. Below, we always assume that image descriptors are vectors from the Euclidean space $\mathbf{R}^D$.

\subsection{Motivation}

All the existing quantization methods contain two essential modules: the \textit{encoder} and the \textit{distance function}. 

The encoder  $f(x) : \mathbf{R}^D \rightarrow \{1,\dots,K\}^{M} $ maps a data vector $x$ into a tuple of $M$ indices $\boldsymbol{i} = \left[i_1,\dots,i_M\right]$. In turn, the distance function $d(q, \boldsymbol{i}) : \mathbf{R}^D \times \{1,\dots,K\}^{M} \rightarrow \mathbf{R}$ estimates how far a query $q$ is from an encoded database vector $f(x)$. Both $f(\cdot)$ and $d(\cdot,\cdot)$ typically depend on learnable parameters (e.g. the quantization codebooks in PQ or the rotation matrix in OPQ).

Currently, the state-of-the-art unsupervised quantization methods use shallow encoders and distance functions. In this study, we instead propose to use deep parametric models for both $f(\cdot)$ and $d(\cdot,\cdot)$ that are jointly trained to perform nearest neighbor retrieval.


\subsection{Model}
\label{sect:model}

The architecture of our model is schematically presented on \fig{arch}.
There are two main parts: the encoder maps the data vector $x$ into a tuple of discrete codes, and the decoder reconstructs the original vector from its compressed representation.

In the encoder part, we use a simple feedforward neural network $net(x)$ 
with $M$ output ''heads'', which are trained jointly. As will be shown below, one can think of $net(x) = \left[net(x)_1,\dots,net(x)_M\right]$ as a mapping of data vectors into a product of $M$ learned spaces. Each of the spaces in this product posesses a codebook of $K$ codewords and we denote by $c_{mk}$ the codeword $k$ from the codebook $m$, $m\in\{1,\dots,M\}, k\in\{1,\dots,K\}$.


We use this model to define the stochastic encoding function that assigns data vectors to discrete codes based on the dot product in the learned space. In particular, the probability of being encoded by the $k$-th code from $m$-th codebook is defined as:
\begin{equation}\label{eq:p_code}
p(c_{mk} | x) = \frac{\exp\left({\langle net(x)_m, c_{mk} \rangle /\tau_m}\right)}{\underset{j}{\sum} \exp\left({\langle net(x)_m, c_{mj}\rangle /\tau_m}\right)} 
\end{equation}

Here $\tau_m \in (0, \infty)$ defines the temperature (inverse "peakyness") of the probability distribution. We treat both $c_{mk}$ and $\tau_m$ as regular model parameters and optimize them with backpropagation.

Assuming conditional independence between the probabilities of codewords in different codebooks for a given $x$, we get:
\begin{equation}\label{eq:p_codes}
p(c_1, ..., c_M | x) = \prod_{i=m}^M p(c_m | x)
\end{equation}

We can now define our encoding function by maximizing over those probabilities:
\begin{equation}\label{eq:p_codes}
\begin{gathered}
f(x) = \underset{c_1, ..., c_M}{\text{argmax }}{ p(c_1, ..., c_M| x)} = \\
= [\underset{c_{1k}}{\text{argmax }}{ p(c_{0k}| x)}, ..., \underset{c_{Mk}}{\text{argmax }}{ p(c_{Mk}| x)}] = \\
= [\underset{c_{1k}}{\text{argmax }}{\langle net(x)_1, c_{1k}\rangle}, ..., \underset{c_{Mk}}{\text{argmax }}{ \langle net(x)_M, c_{Mk}\rangle}]
\end{gathered}
\end{equation}

However, in order to train our model we require encoder to be differentiable. Inspired by\cite{shu}, we use a differentiable approximation of $f(x)$ using Gumbel-Softmax\cite{gumbel, concrete} trick. The differentiable approximation $\tilde{f}_{m}(x)$ for $m$-th codebook is a stochastic function that maps $x$ to a vector:



\begin{equation}\label{eq:gumbel}
    \tilde{f}(x)_{m} = softmax\{\log p(c_{mj}| x) + z_j, j{=}1..K\}
\end{equation}

In the formula above, $z_j$ is a sample from the standard Gumbel distribution that can be obtained as $z_j = - \log(- \log Uniform(0, 1))$. Note that the original Gumbel-Softmax distribution\cite{gumbel} divides all exponent rates by a small temperature factor, which we set to $1$ in all our experiments.

While the relaxation $\tilde{f}(x)_{m}$ is differentiable, it does not produce one-hot vectors, which are needed for quantization methods. Hence during training we discretize $\tilde{f}(x)_{m}$ via using $argmax$ instead of $softmax$ in \eq{gumbel}, which results in one-hot vector, corresponding to the index of the chosen codeword. While this discretization is not differentiable, we backpropagate through it using straight-through gradient estimation: the gradients w.r.t. function outputs are passed to its inputs with no transformation applied.

Then we feed $M$ one-hot vectors, produced by the encoder, into decoder $g(\cdot)$: another feedforward network that adds the corresponding codewords and reconstructs the vector $\tilde x$ in the original data space.

In all our experiments, both $net(\cdot)$ and $g(\cdot)$ are simple fully-connected neural networks with ReLU activation functions and Batch Normalization\cite{BN} layers before each activation (see \fig{arch}). We describe the particular choice of $net(\cdot)$ and $g(\cdot)$ in the experimental section.


\subsection{Nearest Neighbor Search}
\label{sect:inference}

The retrieval of nearest neighbors in a quantized database is performed via the exhaustive search with distance function $d(q, \boldsymbol{i})$:

\begin{equation}
\underset{\boldsymbol{i}}{\text{argmin }} d(q, \boldsymbol{i})
\end{equation}

In our model, we use two different definitions for  $d(q, \boldsymbol{i})$ to provide both high compression accuracy and efficient retrieval.

The first "naive" distance function reconstructs the database vector with the decoder $g$ and measures distance in the original data space:
\begin{equation}\label{eq:d1_reco}
    d_{1}(q, \boldsymbol{i}) = \left\lVert q - g(\boldsymbol{i}) \right \lVert_2^2
\end{equation}

However, the usage of $d_{1}$ for exhaustive search would require applying the decoder network to the whole database, inducing a prohibitively large computational cost for large databases.

Alternatively, one can define the distance function in the learned space of codebooks. This definition relies on the fact that both query mapping $net(q)$ and codewords belong to a shared space. This space is conveniently equipped with a dot-product-based probability \eqref{eq:p_code} of picking a particular codeword. 

Our intuition is that the nearest neighbor of data point $q$ should have codewords that are likely to be assigned to $q$ itself. In other words, in order to search for the nearest neighbors of $q$ we want to consider candidates with the highest $p(c_1, ..., c_M | q)$. This naturally leads us to the following distance function:

\begin{gather}
    d_{2}(q, \boldsymbol{i}) = d_{2}(q, \{i_1,\dots,i_M\}) = - log p(c_{1i_1}, ..., c_{Mi_M} | q) = \nonumber \\ 
    = - \sum_{m=1}^{M} \left( \langle net(q)_m, c_{m{i_m}}  \rangle - \log \sum_{k=1}^{K} \exp \langle net(q)_m, c_{mk}  \rangle \right) = \nonumber \\
    = - \sum_{m=1}^{M} \langle net(q)_m, c_{m{i_m}}  \rangle + const(q)
\label{eq:d2_learned}
\end{gather}


Compared to \eq{d1_reco}, the second formulation allows for an efficient search algorithm via lookup tables. First, the algorithm computes the dot products of $net(q)$ with the codewords in all codebooks using one pass of the encoder network and $O(M \cdot K)$ dot product computations. The algorithm can then find the nearest neighbor by iterating over the encoded data points and summing the cached distances, doing only $M$ additions per database vector. 

The search based on distance function \eq{d2_learned} can be seen as a generalization of existing quantization methods. However, unlike \cite{OPQ,LSQ}, the distance is computed not in the original data space but in a new learned space that is obtained via SGD training.

In practice, we combine both distance functions in a two-stage search: at first, we efficiently select $L$ nearest candidates based on $d_2$, and then re-rank those candidates using the more expensive $d_1$. The additional reranking stage does not influence the total scheme efficiency by much, as only a small fraction of the database is reranked.

Of course, the existing shallow methods can also benefit from the additional reranking with DNN and we compare our scheme with this baseline in the experiments. Our experiments below demonstrate that the post-search reranking slightly increases the accuracy of shallow MCQ methods, but the overall performance of UNQ is substantially higher.

\subsection{Training}
\label{sect:training}

We train our model by explicitly fitting the two distance functions to maximize recall. The first distance function is defined in the original data space and can be trained with autoencoder-like objective:
\begin{equation}\label{eq:loss_sim1}
  L_{1} = \frac{1}{n} \sum_{x_i} d_1(x_i, \tilde{x_i}) = \frac{1}{n} \sum_{x_i} \left\lVert x_i - g(\tilde{f}(x_i)) \right \lVert _2^2
\end{equation}

However, there is no guarantee that minimizing this objective would result in good candidates being selected for the reranking stage. Therefore, we also need to train $d_2(\cdot, \cdot)$ with another objective term.

We employ a metric learning approach by minimizing the triplet loss in the learned space. Intuitively, we want $x$ to be closer to it's true nearest neighbor $x_{+}$ than to the negative example $x_{-}$.

\begin{equation}\label{eq:loss_sim2}
  L_{2} = \frac{1}{n} \sum_{x} max(0, \delta + d_2(x, f(x_{+})) - d_2(x, f(x_{-})))
\end{equation}

Similarly to \cite{catalyst}, we sample $x_{+}$ from top-$3$ true nearest neighbors of data point $x$. In turn, $x_{-}$ is sampled uniformly from between $100$-th to $200$-th nearest neighbors, excluding $x$ itself and three candidates for $x_{+}$. Following a popular practice from the metric learning field, we sample those vectors at the offset of each training epoch.

The final term for our objective is a regularizer for Gumbel-Softmax that encourages equal frequency of codewords. A common problem of nearly all methods for learning discrete variables is that they converge to poor local optima where some codes are (almost) never used. In order to alleviate this issue, we repurpose the squared Coefficient of Variation regularizer that was originally proposed in \cite{ShazeerMMDLHD17} to combat a similar imbalance in the Mixture of Experts layers.

The coefficient of variation is computed from codeword probabilities averaged over the training batch $X$:
$$p_{avg}(i_m | X) = {\frac{1}{n}} \sum_{x_i \in X} p(i_m | x_i)$$

\begin{equation}\label{eq:cvsquared}
CV^2(i_m) = \frac{Var[p_{avg}(i_m | X)]}{\left[E[p_{avg}(i_m |X)]\right] ^ 2}
\end{equation}

Our final objective is just a sum of those three terms with coefficients. In our experiments we pick $\alpha$ from $\{0.1, 0.01, 0.001\}$ via grid search. As for $\beta$, we decrease it linearly from $1.0$ to $0.05$ during training.
\begin{equation}\label{full_objective}
    L = L_1 + \alpha \cdot L_2 + \beta \cdot \frac1M \sum_{m=1}^M CV^2(i_m)
\end{equation}

The model is trained to minimize the training objective $L$ using minibatch gradient descent with the recent Quasi-Hyperbolic Adam algorithm \cite{qhadam}. We also use One Cycle learning rate schedule\cite{smith2017super} for faster model convergence.


\section{Experiments}
\label{sect:experiments}
In this section we provide the experimental results that compare the proposed Unsupervised Neural Quantization (UNQ) method with the existing unsupervised compression methods. Following the recent work\cite{catalyst}, we perform the most of experiments on two sets of data:
\begin{enumerate}
\item \textbf{Deep1M/Deep10M/Deep1B} datasets contain $96$-dimensional visual descriptors, which are computed from the activations of a deep neural network\cite{Babenko16}. Base sets include $10^6$, $10^7$ and $10^9$ vectors correspondingly. We use the additional separate sets of 500.000 vectors for training and 10.000 hold-out queries for evaluation.
\item \textbf{BigANN1M/BigANN10M/BigANN1B} datasets contain $128$-dimensional histogram-based SIFT descriptors\cite{Jegou11b}. Base sets include $10^6$, $10^7$ and $10^9$ vectors correspondingly. Here we also use the separate sets of 500.000 vectors for training and 10.000 hold-out queries for evaluation.
\end{enumerate}

Unless stated otherwise, we always learn the method parameters on the train set, then compress the base set and evaluate the retrieval performance on the query set. As a common measure of compressed-domain retrieval performance we report \textit{Recall@k} (for $k=1,10,100$), which is the probability that the true nearest neighbor is among $k$ closest neighbors in the compressed dataset. Two compression levels ($8$, $16$ bytes per vector) were evaluated. In all the experiments we used the quantization codebooks of $K=256$ codewords for all the methods.
\begin{table*}
\centering
\addtolength{\tabcolsep}{2pt}
\renewcommand\arraystretch{1.1}
\begin{tabular}{|c|ccc|ccc|}
\hline
\multirow{2}{*}{{\centering Method}} & \multicolumn{3}{|c|}{\bf BigANN1M} & \multicolumn{3}{|c|}{\bf Deep1M}\\
 & R@1 & R@10 & R@100 & R@1 & R@10 & R@100\\
\hline
 & \multicolumn{6}{|c|}{\bf 8 bytes per vector}\\
\hline
OPQ & 20.8 & 64.3 & 95.3 & 15.9 & 51.3 & 88.6\\
\hline
Catalyst + OPQ & 26.2 & 73.0 & 97.3 & 20.9 & 61.5 & 93.5\\
\hline
Catalyst + Lattice & 28.9 & 75.8 & 97.9 & 24.6 & 68.3 & 96.1\\
\hline
LSQ & 29.2 & 77.7 & 98.7 & 21.7 & 64.0 & 94.5\\
\hline
LSQ + rerank & 30.3 & 78.9 & 98.8 & 22.8 & 65.7 & 95.6\\
\hline
UNQ & \textbf{34.6} & \textbf{82.8} & \textbf{99.0} & \textbf{26.7} & \textbf{72.6} & \textbf{97.3}\\
\hline
 & \multicolumn{6}{|c|}{\bf 16 bytes per vector}\\
\hline
OPQ & 40.9 & 89.8 & 99.9 & 35.0 & 82.5 & 99.1\\
\hline
Catalyst + OPQ & 46.1 & 92.0 & 99.8 & 39.0 & 86.5 & 99.3\\
\hline
Catalyst + Lattice & 49.1 & 94.1 & \textbf{100.0} & 44.8 & 90.8 & \textbf{99.8}\\
\hline
LSQ & 57.1 & 97.5 & \textbf{100.0} & 41.1 & 88.6 & 99.5\\
\hline
LSQ + rerank & 57.7 & 97.6 & \textbf{100.0} & 42.4 & 89.5 & 99.6\\
\hline
UNQ & \textbf{59.3} & \textbf{98.0} & \textbf{100.0} & \textbf{47.9} & \textbf{93.0} & \textbf{99.8}\\
\hline
\end{tabular}
\vspace{3mm}
\caption{The compressed-domain retrieval performance achieved by unsupervised compression approaches. The proposed UNQ method outperforms all the competitors on both datasets and under both memory budgets.}
\label{tab:main}
\end{table*}
\subsection{Comparison to the state-of-the-art}

As a preliminary experiment, we compare the proposed UNQ method with the current state-of-the-art approaches for the unsupervised compression problem on the million-scale \textbf{Deep1M} and \textbf{Bigann1M} datasets. In particular, we compare the following methods:
\begin{itemize}
    \item \textbf{OPQ}\cite{OPQ,Norouzi13}, with the implementation from the Faiss library\cite{JDH17}.
    \item \textbf{Catalyst{+}OPQ}\cite{catalyst} that uses OPQ on top of the "spreaded" vectors, as described in \cite{catalyst}. We use the implementation provided by the authors and tune $d_{out}$ and $\lambda$ hyperparameters for optimal performance. 
    \item \textbf{Catalyst{+}Lattice}\cite{catalyst} that uses the fixed predefined lattice on a sphere for vector quantization. Here we also use the implementation provided by the authors and tuned $d_{out}$ and $\lambda$ hyperparameters. The dimension of hidden layers was set to $2048$ neurons. The lattice quantizers with $r^2=79$ for $8$ bytes and $r^2=253$ for $16$ bytes were used. 
    \item \textbf{LSQ}\cite{LSQ}, the state-of-the-art shallow quantization method that approximates each vector by a sum of codewords. We use the implementation provided by the authors.
    \item \textbf{LSQ{+}rerank}, that additionally reranks some top of LSQ results by the learned decoder with two hidden layers of $1024$ neurons. The decoder obtains $D$-dimensional LSQ approximations as an input and is trained to minimize the reconstruction objective \eq{loss_sim1}. The number of elements to rerank is the same as for UNQ.
    \item \textbf{UNQ}, our method, introduced in this paper. We use the architecture similar to \cite{catalyst}: the encoder and decoder consist of two 1024-unit linear layers, each equipped with Batch Normalization and ReLU activation function. The dimensionality of codewords was set to $256$, and we rerank top-500 candidates.

\end{itemize}

The recall values achieved by the different methods are presented in \tab{main}. Below we highlight several key observations:
\begin{table*}
\centering
\addtolength{\tabcolsep}{2pt}
\renewcommand\arraystretch{1.15}
\begin{tabular}{|c|ccc|ccc|}
\hline
\multirow{2}{*}{{\centering Method}} & \multicolumn{3}{|c|}{\bf BigANN10M} & \multicolumn{3}{|c|}{\bf Deep10M}\\
 & R@1 & R@10 & R@100 & R@1 & R@10 & R@100\\
\hline
 & \multicolumn{6}{|c|}{\bf 8 bytes per vector}\\
\hline
Catalyst + Lattice & 20.9 & 63.9 & 94.3 & 18.2 & 53.5 & 88.7\\
\hline
LSQ & 21.7 & 64.3 & 95.0 & 14.8 & 48.1 & 84.8\\
\hline
LSQ + rerank & 21.8 & 64.8 & 95.0 & 15.1 & 48.8 & 85.9\\
\hline
UNQ & \textbf{25.8} & \textbf{70.5} & \textbf{96.3} & \textbf{18.8} & \textbf{57.0} & \textbf{90.9}\\
\hline
 & \multicolumn{6}{|c|}{\bf 16 bytes per vector}\\
\hline
Catalyst + Lattice & 42.0 & 90.0 & 99.7 & 37.9 & 84.6 & 99.2\\
\hline
LSQ & 50.5 & 95.0 & \textbf{99.9} & 34.3 & 79.8 & 98.2\\
\hline
LSQ + rerank & 50.7 & 95.3 & \textbf{99.9} & 34.5 & 81.1 & 98.4\\
\hline
UNQ & \textbf{52.1} & \textbf{95.4} & \textbf{99.9} & \textbf{40.1} & \textbf{86.8} & \textbf{99.3}\\
\hline
\end{tabular}
\vspace{3mm}
\caption{The performance on the ten million datasets. On this scale, the advantage of the proposed Unsupervised Neural Quantization method over existing approaches persists.}
\label{tab:10m_table}
\end{table*}
\begin{table*}
\centering
\addtolength{\tabcolsep}{2pt}
\renewcommand\arraystretch{1.15}
\begin{tabular}{|c|ccc|ccc|}
\hline
\multirow{2}{*}{{\centering Method}} & \multicolumn{3}{|c|}{\bf BigANN1B} & \multicolumn{3}{|c|}{\bf Deep1B}\\
 & R@1 & R@10 & R@100 & R@1 & R@10 & R@100\\
\hline
 & \multicolumn{6}{|c|}{\bf 8 bytes per vector}\\
\hline
Catalyst + Lattice & 10.4 & 37.6 & 76.6 & \textbf{16.8} & \textbf{38.7} & 68.2\\
\hline
LSQ & 9.6 & 35.9 & 73.3 & 13.2 & 32.3 & 59.9\\
\hline
LSQ + rerank & 9.9 & 36.1 & 73.8 & 12.3 & 31.6 & 59.7\\
\hline
UNQ & \textbf{13.0} & \textbf{44.5} & \textbf{82.4} & 14.5 & 37.8 & \textbf{68.5}\\
\hline
 & \multicolumn{6}{|c|}{\bf 16 bytes per vector}\\
\hline
Catalyst + Lattice & 31.1 & 77.8 & 98.3 & 35.3 & 72.8 & 95.6\\
\hline
LSQ & 38.0 & 85.6 & 99.3 & 30.5 & 65.0 & 91.1\\
\hline
LSQ + rerank & 37.6 & 86.0 & 99.3 & 30.1 & 65.8 & 91.4\\
\hline
UNQ & \textbf{38.3} & \textbf{86.8} & \textbf{99.4} & \textbf{35.5} & \textbf{74.2} & \textbf{96.1}\\
\hline
\end{tabular}
\vspace{3mm}
\caption{The performance on the one billion datasets. The proposed UNQ method outperforms the competitors in most operating points.}
\label{tab:1b_table}
\end{table*}
\begin{itemize}
    \item On both datasets and for both compression levels the introduced Unsupervised Neural Quantization outperforms the competitors and provides a new state-of-the-art for the unsupervised compression problem.
    \item The current state-of-the-art methods \textbf{Catalyst{+}Lattice} and \textbf{LSQ} are competitive on different types of visual data. While \textbf{LSQ} outperforms 
    \textbf{Catalyst{+}Lattice} on shallow SIFT descriptors, its accuracy is much lower on deep descriptors. Meanwhile, the proposed UNQ provides the highest accuracy on both datasets, which makes it a universal method for all types of data.
    \item An additional reranking stage with a learnable decoder provides only a slight improvement for the shallow \textbf{LSQ} method. This indicates that the end-to-end learning in \textbf{UNQ} is crucial for high compression accuracy.
\end{itemize}

\subsection{Additional memory consumption}

Here we analyze the additional memory consumption required by the proposed UNQ method. Compared to the shallow baselines UNQ additionally stores the parameters of the feed-forward encoder and decoder networks. In our experiments, the model requires about 19.8 Mb for 8-byte budget and 30.1 Mb for 16-byte budget, which is on par with the Catalyst{+}Lattice\cite{catalyst} which requires 17.2 Mb. Note, that this amount does not depend on the number of database vectors. For instance, for databases of one billion vectors, this results only in negligible 0.02 additional bytes per vector. As the most important experiment, we verify that the advantage of UNQ persists for more massive datasets. Namely, we perform the comparison of Catalyst{+}Lattice, LSQ, and UNQ on larger datasets Deep10M and BigANN10M, and billion-scale datasets Deep1B and BigANN1B. The results are shown in \tab{10m_table} and \tab{1b_table} and demonstrate that UNQ always outperforms a shallow LSQ counterpart by a considerable margin. UNQ also outperforms Catalyst{+}Lattice in most operating points, except for R@1,R@10 on 8-bytes encoding on Deep1B. For Deep1B and BigANN1B we rerank top-1000 candidates since it requires a negligible time in comparison with the billion-scale search.




\subsection{Ablation}
\label{sect:ablation}
In this section we empirically validate our choice of training architecture and model by evaluating the contribution of each component. All experiments in this section fit the budget of $M{=}8$ bytes per vector on the BigANN1M dataset. More specifically, we compare

\begin{itemize}
    \item \textbf{UNQ} --- our primary model that is built and trained in accordance with the description provided in Section \ref{sect:method} with all parameters described in the first experiment.
    \item \textbf{Exhaustive reranking} --- like \textbf{UNQ}, but the nearest neighbor search is performed with $d_1(\cdot, \cdot)$ only. This approach requires reconstructing every data vector by running the decoder module once for every vector in the database. This setup was evaluated on the same model parameters as \textbf{UNQ}.
    \item \textbf{No reranking} --- the training procedure is the same as for \textbf{UNQ}, but the search is implemented without reranking by the decoder.
    \item \textbf{No triplet loss} --- UNQ model that performs two-stage search but does not optimize for $d_2(\cdot, \cdot)$ explicitly, i.e. $\alpha=0$. This is equivalent to training a regularized discrete variational autoencoder without explicit requirements to its hidden representation.
    \item \textbf{Triplet only} --- like \textbf{UNQ}, but the nearest neighbor search is performed with $d_2(\cdot, \cdot)$ only. This model is also trained using $\alpha=1.0$ and without the term \eq{loss_sim1}.
    \item \textbf{UNQ w/o hard} --- like \textbf{UNQ}, but Gumbel-Softmax without hard assignment discretization, as in \cite{jang2016categorical}.
    \item \textbf{UNQ w/o Gumbel} --- like \textbf{UNQ}, but the quantization is implemented as in \cite{agustsson2017soft} with $\beta{=}0.1$.
    \item \textbf{No regularizer} --- same as \textbf{UNQ}, but the balance regularizer term is set to $\beta=0$. All other parameters are unchanged.
\end{itemize}

\begin{table}
 \centering
 \renewcommand\arraystretch{1.3}
 \renewcommand\arraystretch{1.1}
\begin{tabular}{|c|ccc|}
\hline
\multirow{2}{*}{{\centering Method}} & \multicolumn{3}{|c|}{\bf BigANN1M, 8 bytes}\\
& R@1 & R@10 & R@100\\
 \hline
 UNQ &  34.6 & 82.8 & 99.0 \\
 \hline
 Exhaustive reranking & 34.6 & 82.8 & 99.3 \\
 \hline
 No reranking & 25.0 & 68.5 & 95.0 \\
 \hline
 No triplet loss & 35.5 & 83.4 & 95.7 \\
 \hline
 Triplet only & 27.9 & 72.6 & 99.2 \\
 \hline
 UNQ w/o hard & 33.8 & 80.4 & 98.0 \\
 \hline
 UNQ w/o Gumbel & 30.2 & 75.7 & 78.1 \\
 \hline
 No regularizer & 31.0 & 80.4 & 95.2 \\
 \hline 
\end{tabular}
\vspace{1mm}
\caption{Ablation study for different training objectives.}
\label{tab:ablation}
\end{table}

The results in \tab{ablation} suggest that each of the three core components of our approach (reranking, triplet loss, CV regularizer) influences the model performance. The reranking stage with $L=500$ candidates is predictably less important on $R@100$, compared to $R@1$, as one can see from the comparison between \textbf{UNQ} and \textbf{No reranking}. The triplet loss term benefits $R@100$, while the CV regularizer provides significant gains across all three recall areas. Note, that the usage of the Gumbel-Softmax trick outperforms the differentiable quantization, proposed in \cite{agustsson2017soft}. Finally, the usage of "hard" version of Gumbel-Softmax trick results in higher performance compared to \textbf{UNQ w/o hard} option.

\subsection{Timings}
Finally, we discuss the timings needed to encode the database and search over compressed data.


\textbf{Encoding} the database points with UNQ has almost the same complexity as in Catalyst as it requires the only feed-forward pass through two fully-connected layers. In particular, the encoding time of Deep1M for $8$ bytes per point on the single Nvidia 1080Ti GPU for the UNQ requires about 1.5 seconds, while for the Catalyst{+}Lattice it is about 4.1 seconds on the same GPU card. The LSQ encoding is slower as it requires several optimization iterations, in our experiments it took 27 seconds to encode Deep1M with the authors' implementation.

\textbf{Search.} Unlike the existing competitors, the proposed UNQ method includes an additional reranking stage, that reconstructs a few candidates with the feed-forward decoder and computes the distances in the original $D$-dimensional space. Because the number of candidates is typically small, the reranking stage almost does not influence the total search runtime. E.g. on the Deep1B dataset with $M{=}8$ the exhaustive scan with $d_2(\cdot,\cdot)$ (via lookup tables) requires 3 seconds. Meanwhile, the reranking of 1000 candidates, implemented via BLAS instructions with the Intel MKL library, requires only 25.9 ms. Both timings are obtained in a single-CPU mode on the same machine. This indicates that the additional runtime cost from reranking is insignificant, especially for large databases or longer codes. Note, that the search in the Catalyst{+}Lattice method is slower compared to the LUT-based methods, namely, \cite{catalyst} reports about $1.5\times$ increase in search runtime for $8$-byte codes.

\section{Conclusion}
\label{sect:conclusion}

We have presented Unsupervised Neural Quantization (UNQ) --- a new unsupervised compression scheme for the problem of compressed-domain retrieval. Our scheme employs the ideas from the recent works on discrete auto-encoders and shows that with the proper training objective the hidden variables can successfully serve as quantized representations for efficient retrieval. From another point of view, our method can be seen as a natural "deep" generalization of the existing shallow quantization methods, such as AQ or LSQ. 

By a large number of experiments, we demonstrate the advantage of UNQ over the state-of-the-art approaches, such as LSQ and the recent lattice-based quantizer. Furthermore, while the existing methods perform differently on different types of data, UNQ provides the highest retrieval accuracy on both histogram-based and deep descriptors. 

For the reproducibility purposes, we publish the Pytorch implementation of Unsupervised Neural Quantization online\footnote{\texttt{https://github.com/stanis-morozov/unq}}.

{\small
\bibliographystyle{ieee_fullname}
\bibliography{egbib}
}

\end{document}